\documentclass{article}

\usepackage{PRIMEarxiv}

\usepackage[utf8]{inputenc} 
\usepackage[T1]{fontenc}    
\usepackage{hyperref}       
\usepackage{url}            
\usepackage{booktabs}       
\usepackage{amsfonts}       
\usepackage{amssymb}        
\usepackage{amsmath}        
\usepackage{nicefrac}       
\usepackage{microtype}      
\usepackage{lipsum}
\usepackage{fancyhdr}       
\usepackage{graphicx}       
\graphicspath{{media/}}     
\usepackage{listings}
\usepackage{xcolor}

\title{OliveGemma: A 3 Billion Visual–Language Model for Recognising the Mediterranean \& European Diet
}

\author{
  Dimitrios I. Zaridis \\
  Unit of Medical Technology \\ 
  \& Intelligent Information Systems \\
  University of Ioannina \\
  Ioannina, Greece \\
  \texttt{dimzaridis@gmail.com} \\
   \And
  Traianos Tsiokris* \\
  Unit of Medical Technology \\
  \& Intelligent Information Systems \\
  University of Ioannina \\
  Ioannina, Greece \\
  \texttt{akistsiokris@gmail.com} \\
     \And
  Vasileios C. Pezoulas \\
  Unit of Medical Technology \\
  \& Intelligent Information Systems \\
  University of Ioannina \\
  Ioannina, Greece \\
  \texttt{bpezoulas@gmail.com} \\
     \And
  Daphni Plati \\
  Unit of Medical Technology \\
  \& Intelligent Information Systems \\
  University of Ioannina \\
  Ioannina, Greece \\
  \texttt{daphni.plati@gmail.com} \\
  \And
    Eugenia Mylona \\
  Unit of Medical Technology \\
  \& Intelligent Information Systems \\
  University of Ioannina \\
  Ioannina, Greece \& \\
  Department of Medical Physics, School of Medicine \\
  University of Patras \\
  Patras, Greece \\
  \texttt{mylona.eugenia@gmail.com} \\
  \And
     Eleni Georga \\
  Unit of Medical Technology \\
  \& Intelligent Information Systems \\
  University of Ioannina \\
  Ioannina, Greece \& \\
  \texttt{egeorga@uoi.gr}\\
  \And
  Nikos Tsiknakis \\
Computational BioMedicine Laboratory \\
Foundation for Research and Technology Hellas \\
Heraklion, Greece \\
\texttt{tsiknakisn@ics.forth.gr}\\
  \And
  Antonis Sakellarios \\
  Dept. of Mechanical and \\
  Aeronautics Engineering \\
  University of Patras \\
  Patras, Greece \& \\
  \texttt{asakellarios@upatras.gr}\\
     \And
  Dimitrios I. Fotiadis \\
  Unit of Medical Technology \\
  \& Intelligent Information Systems \\
  University of Ioannina \\
  Ioannina, Greece \\
  \& Biomedical Research Institute \\
  FORTH \\
  Ioannina, Greece
  \texttt{fotiadis@uoi.gr} \\
}
\begin{document}
\maketitle

\begin{abstract}

Image-based dietary assessment offers a scalable alternative to error-prone self-reported food diaries, yet fine-grained food recognition remains challenging due to high intra-class variability, visually similar dishes, long-tailed distributions, and the compositional nature of plated meals. This study presents OliveGemma, a domain-adapted vision–language model for recognising and reasoning about Mediterranean and European cuisine. Built on the open-weight PaliGemma-2-3B architecture, OliveGemma is fine-tuned with Low-Rank Adaptation (LoRA) on a unified corpus of 17,340 images from three European research project datasets (MedGR, ODIN, and VIPPSTAR), reconciled into a vocabulary of 216 composed dish categories and paired with 102,642 instruction-style question–answer items covering dish recognition, likely and visible ingredients, class-boundary discrimination, visual evidence and overall visual food understanding. Only 23.75M parameters (0.78\% of the 3B backbone) are updated, yielding a $\approx 90$MB adapter deployable on commodity hardware (CPU, 16 GB RAM), being open-source. Under a three-fold cross-validation scheme with an identical closed-vocabulary protocol, OliveGemma achieves a top-1 accuracy of $\overline{92.96} \pm 0.91 \%$, exceeding the strongest CNN baseline (DenseNet-121) by $7.31\%$ and outperforming zero-shot frontier models with exact instructions and bounded classes (same 216 classes as OliveGemma and CNNs) including Gemini Flash 3 and 3.5, GPT-5.4 Mini, and Claude Haiku 4.6 by $\approx 18\%, 46\%, 64\%$, respectively. Furthermore, OliveGemma demonstrates competitive performance on Top-3 and Top-5 accuracy, being the second best across CNNs and Frontier models, only surpassed by DenseNet-121. In addition, OliveGemma achieves $\overline{90.79} \pm 1.3\%$ Exact-Set in Likely Ingredients of the food categories, detecting effectively the visually evident ingredients presented in the dish. These results demonstrate that parameter-efficient adaptation of a small open-weight VLM can surpass substantially larger proprietary systems on specialised food recognition while enabling privacy-preserving, reproducible deployment for dietary assessment. The model is publicly available at \url{https://huggingface.co/JamesZar/OliveGemma-3B}, under the name \textit{JamesZar/OliveGemma-3B}, while the experiments and the results shall be found from the following git repository \url{https://github.com/tsiokris/OliveGemma}.

\end{abstract}

\keywords{Large Language Models; Visual Language Models; Fine tuning; Food Recognition}

\section{Introduction}

Diet is widely recognised as one of the most important modifiable determinants of long-term health. Among healthy dietary patterns, the Mediterranean diet has consistently been associated with a reduced risk of cardiovascular disease, several chronic conditions, and improved longevity \cite{estruch2018predimed}. However, translating this evidence into clinical practice and everyday life requires accurate and practical methods for monitoring dietary intake. Traditional dietary assessment methods, such as self-reported food diaries and dietary recalls, are time consuming and sensitive to reporting and recall bias. Consequently, image based dietary assessment has emerged as a scalable alternative, enabling automatic analysis of meal photographs to identify consumed foods and estimate nutritional intake \cite{wang2022dietary}. Automated food recognition forms the core component of such systems and has applications in clinical nutrition monitoring, consumer food-logging platforms and intelligent food service technologies \cite{min2019survey}.

Despite the progress in computer vision, food recognition remains a challenging task. Unlike many conventional object recognition problems, food images exhibit high intra-class variability, as the same dish may differ substantially in appearance depending on ingredients, preparation methods, plating style, or lighting conditions. On the other hand, visually similar dishes often belong to different categories, resulting in low inter-class variability. Food datasets also exhibit a pronounced long-tail distribution, where a small number of common dishes dominate while many categories contain relatively few examples. An additional challenge arises from the compositional nature of meals, as a single plate frequently contains multiple food components rather than a single homogeneous dish. These challenges become even more pronounced when combining heterogeneous datasets originating from different countries, languages, and annotation protocols, where identical dishes may appear under different names or different levels of semantic granularity.

Early research addressed food recognition as a closed-vocabulary image classification problem, where convolutional neural networks (CNNs) are trained to assign each image to one predefined class. Benchmarks such as Food-101 \cite{bossard2014food101} and later Food2K \cite{min2023food2k} established this paradigm and enabled substantial improvements in recognition accuracy. Nevertheless, conventional image classifiers remain inherently limited by their fixed label space, producing only a single class prediction without providing additional semantic understanding of the image.

Recent advances in vision-language models (VLMs) have introduced a more flexible alternative by combining a pretrained visual encoder with a large language model capable of generating natural language responses. Architectures such as BLIP-2 \cite{li2023blip2}, LLaVA \cite{liu2023llava}, Qwen2.5-VL \cite{bai2025qwenvl}, and PaliGemma-2 \cite{steiner2024paligemma2} extend image recognition beyond fixed classification by supporting instruction following, open ended visual understanding and reasoning about image content. These capabilities are prominent for food analysis, where recognising a dish may also require distinguishing visually similar meals, identifying ingredients, or explaining the visual evidence supporting a prediction.

Despite these advances, several limitations remain. General-purpose VLMs are not specifically trained for fine grained food recognition and often struggle to distinguish visually similar dishes, particularly when operating in specialised domains such as Mediterranean cuisine. In addition, the strongest commercial VLMs are accessible only through proprietary cloud based APIs, raising concerns regarding reproducibility, deployment cost, and privacy, particularly in clinical applications where patient images should remain within institutional infrastructure. Although full fine-tuning can adapt a foundation model to a specialised domain, updating billions of parameters is computationally demanding and increases the risk of catastrophic forgetting \cite{kirkpatrick2017overcoming}. Parameter efficient finetuning methods, particularly Low-Rank Adaptation (LoRA), provide an attractive alternative by updating only a small fraction of the model parameters while preserving the pretrained backbone \cite{hu2022lora,dettmers2023qlora}. However, the application of LoRA-based adaptation to composed, cross-dataset food recognition has received limited attention, and comparisons against both conventional CNN classifiers and contemporary frontier VLMs under identical evaluation conditions remain limited.

Motivated by these limitations, this study presents \textbf{OliveGemma}, a domain adapted vision language model based on the open-weight PaliGemma-2-3B architecture. The model is finetuned using LoRA on a unified corpus of 17,340 food images collected from the MedGR, ODIN, and VIPPSTAR datasets and reconciled into a canonical vocabulary of 216 composed dish categories. By updating only 23.75 million trainable parameters (0.78\% of the approximately three billion parameters of the original model), OliveGemma provides an efficient and reproducible solution for fine-grained food recognition while preserving the advantages of an open-weight deployment. Experimental evaluation demonstrates that the proposed approach achieves competitive recognition performance, outperforming several established CNN architectures and substantially exceeding the zero-shot performance of state of the art proprietary vision language models evaluated under an identical closed vocabulary protocol.

\subsection{Related Work}

\subsubsection{Convolutional Neural Networks Food Image Recognition}

Automated food recognition has traditionally been formulated as a closed vocabulary image classification task and has advanced alongside the development of increasingly large datasets. Food-101 \cite{bossard2014food101} established the standard benchmark for 101 Western dishes, while the UEC-Food datasets expanded the problem to Japanese cuisine \cite{kawano2014uecfood256}. Larger datasets, including ISIA Food-500 \cite{min2020isia} and Food2K \cite{min2023food2k}, further increased the label space to 500 and 2,000 food categories, respectively. Complementary resources such as Recipe1M \cite{salvador2017recipe1m} introduced paired recipe-image supervision, enabling multimodal learning for food understanding. Furthermore, especially for Mediterranean diet, there is the MedGR dataset, used in this analysis, with over 50K images depicting food plates of greek-italian cuisine \cite{medgr}. Despite their scale and diversity, these datasets share a common assumption that each image is assigned to a single predefined class within a flat taxonomy. Consequently, they do not support composed meals containing multiple food components, nor do they address the reconciliation of heterogeneous food vocabularies originating from different countries.

On the modeling side, CNNs have been the dominant approach for food image classification. Architectures such as ResNet \cite{he2016resnet}, Inception (GoogLeNet) \cite{szegedy2015googlenet}, DenseNet \cite{huang2017densenet}, and EfficientNet \cite{tan2019efficientnet} have consistently demonstrated strong recognition performance, while more recent studies have adopted Vision Transformers as image classifiers \cite{dosovitskiy2021vit}. Given the fine-grained nature of food recognition, several works have incorporated attention mechanisms and second-order or bilinear pooling techniques to capture subtle visual cues that distinguish highly similar dishes \cite{lin2015bilinear}. Although these methods achieve high accuracy within a fixed taxonomy, they remain inherently limited by the closed-vocabulary classification paradigm. Their predictions are restricted to predefined labels and they cannot explain or justify their decisions.

\subsubsection{Vision Language Models and Instruction Tuning}

Recent advances in VLMs have shifted visual recognition from fixed-label classification toward open-ended language generation. This stack is built upon contrastively pretrained vision encoders, most notably CLIP \cite{radford2021clip} and its successor SigLIP \cite{zhai2023siglip}, which align images and text within a shared embedding space. Generative VLMs combine these visual encoders with large language models to enable multimodal reasoning and natural language responses. BLIP-2 \cite{li2023blip2} bridges the visual and language components through a lightweight querying transformer, whereas instruction-tuned models such as LLaVA \cite{liu2023llava}, the Qwen-VL family \cite{bai2023qwenvl,bai2025qwenvl}, InternVL \cite{chen2024internvl}, PaliGemma and PaliGemma-2 \cite{beyer2024paligemma,steiner2024paligemma2} extend this framework by following natural language instructions and reasoning over visual content.

Among these architectures, PaliGemma-2 provides several characteristics that make it suitable for domain adaptation because it combines a SigLIP vision encoder with a Gemma-2 language decoder through a lightweight linear projection layer and is released as an open-weight model at a practical 3B parameter scale. Unlike chat-oriented assistants, PaliGemma-2 is designed as a transferable vision language foundation model and achieves strong performance across a wide range of downstream vision language tasks after task-specific finetuning \cite{steiner2024paligemma2,beyer2024paligemma}. These characteristics make it suitable  backbone candidate for developing reproducible and locally deployable food recognition systems.

Despite the rapid progress of VLMs, important challenges remain. General purpose models frequently underperform in specialised fine-grained domains when evaluated in the zero-shot setting, particularly when subtle visual differences separate food categories. Furthermore, many state-of-the-art commercial VLMs are accessible only through proprietary cloud APIs, limiting reproducibility while introducing deployment costs and privacy concerns. These limitations are especially relevant in clinical dietary assessment, where patient images may contain protected health information and therefore processing of those data may be prohibited by external services.

\subsubsection{Parameter-Efficient Fine-Tuning and Domain Adaptation}
Fully fine-tuning a multi billion parameter VLM is computationally expensive and risks catastrophic forgetting of the backbone's general capabilities \cite{kirkpatrick2017overcoming}. On the other hand parameter-efficient finetuning (PEFT) 
updates only a small set of added or selected parameters while the backbone stays frozen. The PEFT family includes bottleneck adapters \cite{houlsby2019adapter}, prefix and prompt-tuning \cite{li2021prefix,lester2021prompt}, and Low-Rank Adaptation (LoRA) \cite{hu2022lora}, grounded in the hypothesis that the weight update required to adapt a pretrained model has low intrinsic rank, together with its quantised variant QLoRA \cite{dettmers2023qlora}. 

For multimodal foundation models, PEFT is frequently combined with staged adaptation strategies in which the language model is first aligned to the target task before selectively unfreezing components of the vision encoder. This approach has become a common strategy for transforming general purpose VLMs into domain-specific models. For instance, LLaVA-Med adapts LLaVA for biomedical image understanding \cite{li2023llavamed}, and GeoChat, which applies LoRA-based fine-tuning to remote sensing imagery while demonstrating that low-rank adaptation preserves the backbone model's general capabilities \cite{kuckreja2024geochat}.

OliveGemma follows the same general pattern by adapting an open-weight vision-language model to the food domain using LoRA. However, to the best of our knowledge, no previous study has applied this strategy to composed label food recognition across multiple heterogeneous datasets, nor evaluated the resulting model against both conventionally trained CNN classifiers and zero-shot frontier VLMs using an identical dataset, label space, and evaluation protocol.

\subsection{Contributions}
The main contributions of this study are summarized as follows.

\begin{itemize}

\item We propose an open-weight vision-language model, \textbf{OliveGemma}, based on PaliGemma-2-3B, adapted using LoRA for fine-grained composed food recognition. The model is trained on a unified dataset of 17,340 images compiled from the MedGR, ODIN, and VIPPSTAR datasets and reconciled into a canonical vocabulary of 216 food categories.
\item We perform a comparative evaluation against both conventional image classification models and existing vision-language models. The benchmark includes fine-tuned CNN architectures, OliveGemma, and zero-shot frontier VLMs from Google (Gemini), OpenAI (GPT), and Anthropic (Claude), all evaluated using the same dataset split and label space to isolate the effect of domain adaptation.
\item We extend food recognition beyond single-label classification through multimodal reasoning capabilities, including ingredient prediction and natural language generation of visual evidence and visible ingredients, supporting richer dietary assessment and explainable food analysis.
\item OliveGemma can be deployed on local premises and can operate on CPU with 16GB of RAM, enabling privacy-preserving operalization.

\end{itemize}

\section{Methodology}

\subsection{Overall Workflow}
\begin{figure}
    \centering
    \includegraphics[width=1 \linewidth]{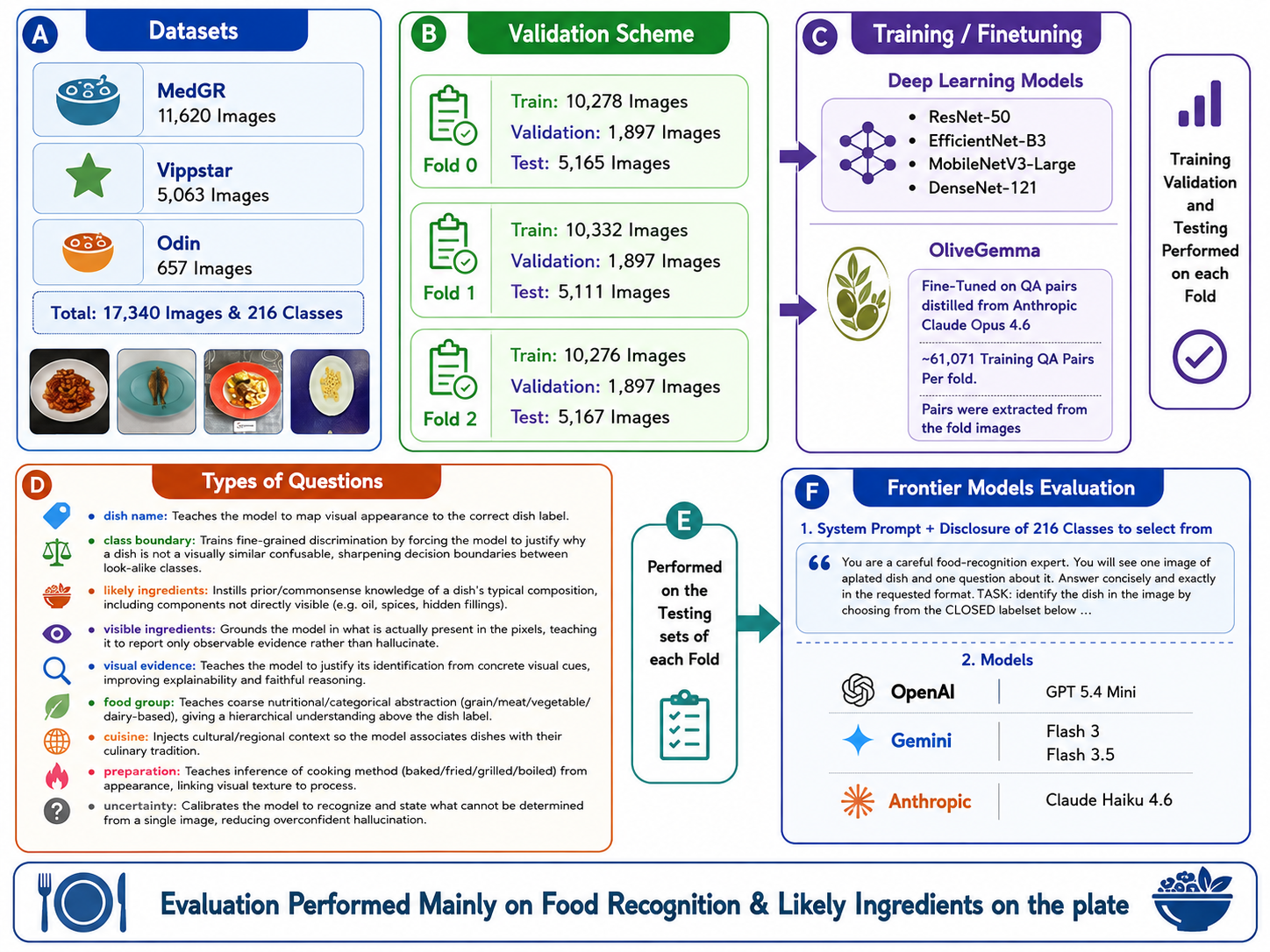}
    \caption{The overall workflow of this study}
    \label{fig:workflow}
\end{figure}

Figure \ref{fig:workflow} illustrates the overall workflow followed in this study. Initially, the MedGR, ODIN, and VIPPSTAR datasets were collected, harmonized, and structured to support a unified experimental pipeline. A 3 fold cross validation scheme was then applied while preserving the original class imbalance and ensuring sufficient representation of all food categories across the three folds. For the CNN baseline experiments, four established image classification architectures pretrained on the Food-101 dataset were selected to leverage prior domain knowledge of food images. Each model was finetuned and evaluated independently on every fold.

The proposed OliveGemma model was fine-tuned using question-answer (QA) pairs generated from the annotations already available in the employed datasets together with knowledge distilled from Anthropic Claude 4.6 Opus. Approximately 61,071 QA pairs were generated for each fold. The training corpus was designed to capture multiple aspects of food understanding, including food recognition, visual reasoning between visually similar dishes, likely ingredients, visual evidence supporting the predicted class, food groups, cuisine identification, preparation steps, and uncertainty-aware responses. The latter encourages the model to explicitly acknowledge information that cannot be reliably inferred from a single image, reducing overconfident hallucinations during inference.

To compare OliveGemma against frontier vision-language models, OpenAI GPT-5.4 Mini, Google Gemini Flash 3 and Gemini Flash 3.5, and Anthropic Claude Haiku 4.6 were evaluated. To ensure a consistent evaluation protocol, all models were prompted using the same system prompt, instructing them to act as food recognition experts. Furthermore, the complete canonical class vocabulary was provided, and each model was constrained to return exactly one food category from the predefined label set, thereby isolating the effect of model capability from differences in prompting or output formatting.

\subsection{Dataset Description}

The dataset used in this study was constructed by merging three heterogeneous
datasets, MedGR, ODIN, and VIPPSTAR each originating from a separate European research project. Combining them produced a single image collection $\mathcal{D}=\{(I_i,\,s_i,\,d_i)\}_{i=1}^{N}$ of $N=17{,}340$ images, where $s_i$ denotes the source and $d_i$ the dish label. Across the raw material there were 250 distinct dish folders, but many of these described the same food under different languages or different levels of detail. For example, the Italian pollo and the English chicken refer to an identical dish yet appear as separate folders.  To remove this redundancy, the source folders were reconciled into one canonical recognition vocabulary $\mathcal{C}$ with $|\mathcal{C}|=216$ composed dish labels, listed in appendix (Table \ref{tab:food_paragraph}).  Every canonical label describes a complete plate and may itself comprise several components, so the recognition gold standard $d_i\in\mathcal{C}$ assigned to each image corresponds to this whole-plate canonical label. Every image was converted to
RGB, min–max normalised to $[0,255]$ and resized to the $448\times448$ input expected by the SigLIP encoder.

Each image is paired with one or more instruction-style question–answer items. The dish label itself was taken directly from the original dataset annotations, whereas the remaining contextual information about ingredients was distilled using Claude Opus 4.6 as a teacher model. Five attribute types relevant to recognition and fine-grained reasoning were retained for fine-tuning: the dish name (the food recognition task), the likely ingredients, the class boundary that separates a dish from a similar one, the visible ingredients, and the supporting visual evidence. This procedure yielded $102{,}642$ QA pairs in total.  Each training instance takes the form of a triplet $(I, x, y)$ comprising the image, the prompt $x$ and the target answer $y$, with the dish-name recognition task covering all $216$ canonical classes of $\mathcal{C}$. Recognition supervision was available only for the $17{,}340$ images carrying a canonical dish-name answer, the remaining images contributing solely to the auxiliary attributes. These $17{,}340$ images formed the recognition evaluation set used throughout the cross-model comparisons, and their assignment to folds was held fixed.

\subsection{OliveGemma Fine Tuning Strategy}

The proposed model adapts the open-weight vision-language model PaliGemma-2-3B to the task of fine-grained food recognition and attribute reasoning. PaliGemma combines a SigLIP vision encoder with a Gemma-2 autoregressive language decoder through a linear projection layer that maps visual embeddings into the language model's token space. Given an input image $I$ of size $448\times448$ and a textual prompt $x$, the vision encoder produces $1{,}024$ visual tokens that are prepended to the tokenized prompt. The decoder subsequently generates the target response $y=(y_1,\dots,y_T)$ in an autoregressive manner.

Training follows the standard supervised objective used by decoder-only language models. Let $\Theta$ denote the complete set of model parameters. The optimization objective is the next-token cross-entropy loss, computed only over the target answer tokens, while the image and prompt tokens are excluded from the loss calculation:

$$
\mathcal{L}(\Theta) \;=\; -\sum_{t=1}^{T} \log p_{\Theta}\!\left(y_t \,\middle|\, y_{<t},\, I,\, x\right).
$$

Here, $p_{\Theta}(y_t \mid y_{<t},I,x)$ denotes the probability assigned to the next token given the input image, textual prompt, and previously generated tokens. The sequence $y_{<t}$ represents all target tokens preceding position $t$, and $T$ denotes the length of the target response.

Rather than updating all approximately $3.03\times10^9$ model parameters, parameter-efficient fine-tuning is performed using Low-Rank Adaptation (LoRA). This approach freezes the pretrained backbone and learns only a small set of low-rank weight updates, reducing memory requirements while mitigating catastrophic forgetting.

For every adapted projection matrix $W_0\in\mathbb{R}^{d\times k}$, the forward computation becomes

$$
h \;=\; W_0\,z \;+\; \Delta W\,z \;=\; W_0\,z \;+\; \frac{\alpha}{r}\,B A\,z,
\qquad B\in\mathbb{R}^{d\times r},\; A\in\mathbb{R}^{r\times k},\; r\ll \min(d,k),
$$

In this formulation, $W_0$ denotes the frozen pretrained weight matrix, $z$ is the input activation, and $h$ is the corresponding output activation. The trainable matrices $A$ and $B$ constitute the LoRA adaptation, whose product has rank at most $r$. Consequently, the number of trainable parameters is reduced from $dk$ to $r(d+k)$ for each adapted projection.

The LoRA configuration follows the standard parameterization with rank $r=16$, scaling factor $\alpha=32$, corresponding to an effective scaling of $\alpha/r=2$, and a dropout rate of 0.05 applied to the low-rank branch. Matrix $A$ is initialized from a zero-mean Gaussian distribution $\mathcal{N}(0,\sigma^2)$, while $B$ is initialized to zero. Consequently, $\Delta W=0$ at initialization, ensuring that optimization starts exactly from the pretrained PaliGemma-2 model. Throughout fine-tuning, only the LoRA parameters ${A,B}$ are updated, whereas the vision encoder, projection layer, and original language model weights remain frozen.

\subsubsection{Training \& Frozen Components}

\begin{figure}[hbpt]
    \centering
    \includegraphics[width=1\linewidth]{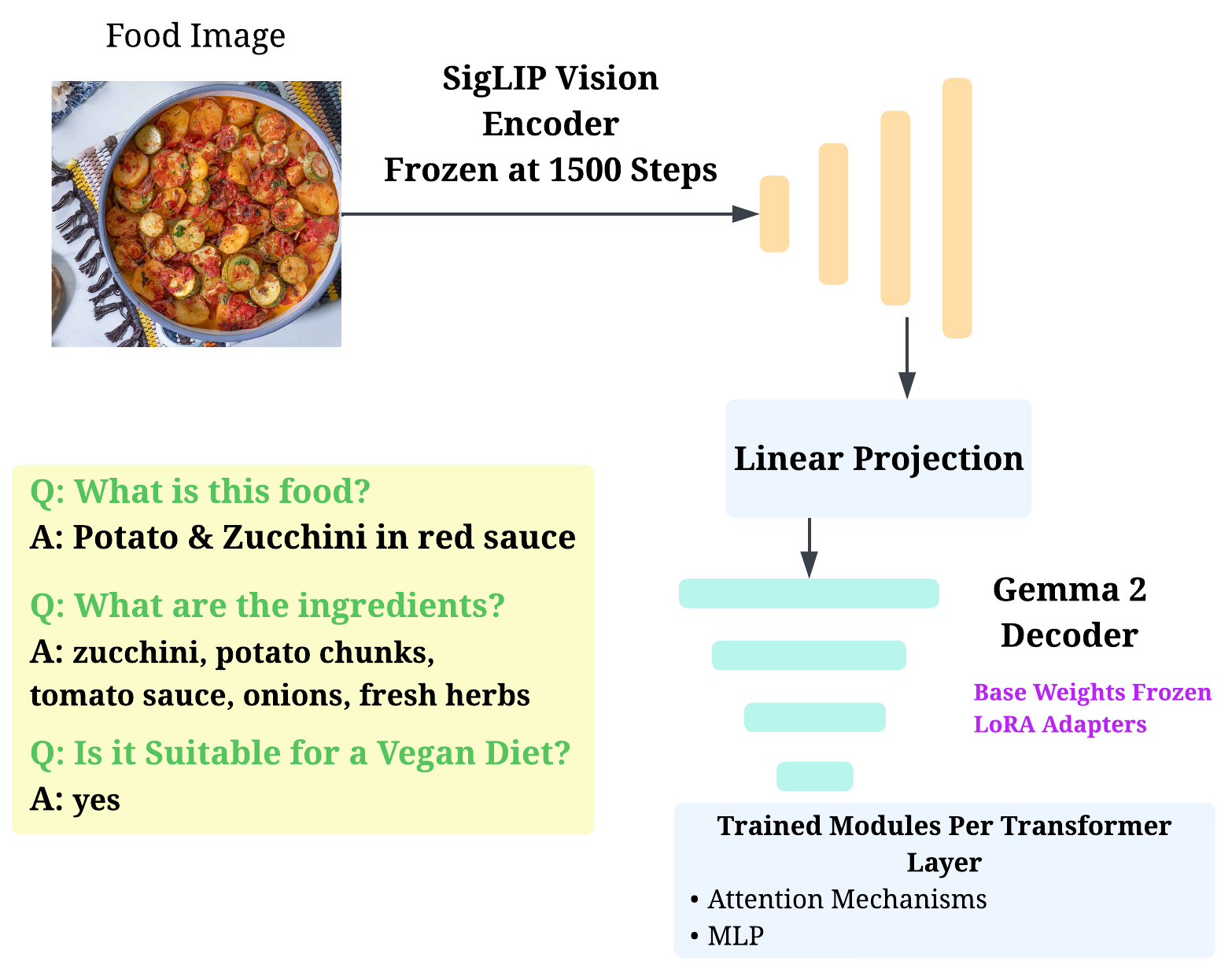}
    \caption{Parameter-efficient fine-tuning of PaliGemma-2-3B. The food image
is encoded by SigLIP (frozen for 1,500 steps, then fine-tuned at $0.1\times$ the
base learning rate), projected into the decoder token space (frozen projector and
embeddings), and decoded by Gemma-2. Only rank-16 LoRA adapters on the attention
$\{q,k,v,o\}$ and MLP $\{\text{gate},\text{up},\text{down}\}$ projections are
trained, therefore the effective training parameter kept at $23.75$ M ($0.78\%$ of the $3$ B backbone).}
    \label{fig:lora_meth}
\end{figure}

LoRA modules are inserted into seven projection layers of every Gemma-2 decoder block which are the four self-attention projections ${W_q, W_k, W_v, W_o}$ and the three feed-forward (SwiGLU) projections ${W_{\text{gate}}, W_{\text{up}}, W_{\text{down}}}$. Figure~\ref{fig:lora_meth} illustrates the resulting architecture together with the decomposition of trainable and frozen components. The token embedding layer, multimodal projection layer, and language modeling head remain frozen throughout training. The SigLIP vision encoder follows a two-stage optimization schedule designed to first align the language decoder with the target output format before adapting the visual representations. During the first $S=1{,}500$ optimization steps, the entire vision encoder is frozen, such that $\nabla_{\theta_v}\mathcal{L}\equiv0$. After this warm-up phase, the encoder is unfrozen and optimized jointly with the decoder using a learning rate reduced by a factor of ten, i.e., $\eta_v=\rho\eta$ with $\rho=0.1$.

This strategy results in only 23,752,704 trainable parameters, corresponding to approximately 0.78\% of the full PaliGemma-2 backbone (3.03 billion parameters). The resulting LoRA adapter occupies approximately 90 MB of storage, enabling efficient training and deployment while preserving the pretrained model weights.
$$
|\theta_{\text{train}}| \;=\; \underbrace{23{,}752{,}704}_{\text{LoRA}} \;\approx\; 0.78\%\ \text{of}\ |\Theta|,
$$

\subsection{CNN Models}

To establish conventional closed-vocabulary baselines, four representative CNN architectures were selected, namely, the ResNet-50 \cite{he2016resnet}, the EfficientNet-B3 \cite{tan2019efficientnet}, the MobileNet-V3-Large \cite{howard2019mobilenetv3}, and the DenseNet-121 \cite{huang2017densenet}. These models represent the principal families of image classification architectures while spanning different trade-offs between accuracy, computational cost, and parameter efficiency. ResNet-50 serves as the reference residual architecture, employing bottleneck residual blocks to facilitate stable optimization and producing a 2,048-dimensional pooled feature representation. EfficientNet-B3 represents the family of compound scaled networks, jointly scaling depth, width, and input resolution to achieve a favorable accuracy to computation trade off, offering 1536 embeddings. MobileNet-V3-Large combines inverted residual blocks, depthwise separable convolutions, squeeze-and-excitation modules, and hard-swish activations to provide an architecture optimized for mobile and edge devices, producing 1,280 image embeddings. DenseNet-121 employs densely connected convolutional blocks that promote feature reuse and efficient gradient propagation, generating a 1,024 embeddings that has proven effective for fine grained visual recognition tasks. To ensure that performance differences are attributable to the backbone architecture rather than the training procedure, all models follow an identical transfer learning protocol. Networks are initialized using the ImageNet-1K pretrained weights, while the convolutional feature extractor remains frozen throughout training. Only the final linear classification layer is optimized, mapping the pooled feature representation to the 216 food categories. Consequently, each model updates between approximately 255,000 and 510,000 trainable parameters. Input images are resized to $224\times224$ pixels and normalized using the standard ImageNet mean and standard deviation. Data augmentation consists of random resized cropping and random horizontal flipping during training. Optimization is performed using Adam with a learning rate of $1\times10^{-3}$, a batch size of 32, and a training duration of 10 epochs. The checkpoint achieving the lowest validation loss is retained for evaluation. Following the evaluation protocol described in subsection \ref{evaluation}, all CNN baselines are trained and evaluated using the same three-fold cross-validation splits employed for OliveGemma. Performance is reported using top-1, top-3, and top-5 accuracy on the held-out test fold, ensuring a consistent comparison across all evaluated models.

\subsection{Evaluation} \label{evaluation}

The evaluation protocol was designed to ensure a consistent and fair comparison across all modelling approaches, including the CNN baselines, OliveGemma, and the frontier VLMs. A three-fold cross-validation scheme was adopted, where in each iteration one fold was used for training, one for validation, and the remaining fold for testing. The assignment of samples to each fold remained fixed throughout all experiments, ensuring that every model was evaluated on exactly the same training, validation, and test splits. To enable a fair comparison with frontier VLMs, the same evaluation protocol was applied to the Gemini, GPT, and Claude model families. All models received an identical system prompt instructing them to act as food recognition experts. Furthermore, the complete canonical class vocabulary was provided, and the models were constrained to predict only a food category belonging to this predefined label set. This converted the otherwise open vocabulary models into a closed-vocabulary classification setting directly comparable with the CNN baselines and OliveGemma. Model performance was evaluated using top-1, top-3, and top-5 accuracy. Given the relatively large number of food categories and the visual similarity among many dishes, these metrics provide a more informative assessment than top-1 accuracy alone by accounting for correct predictions appearing among the highest-ranked candidate classes.

\section{Results}

\subsection{Comparison with CNN Baselines}

\begin{table}[htbp]
  \caption{Performance Evaluation across the 3-fold Cross-Validation Scheme between CNN models and OliveGemma}
  \centering
  \begin{tabular}{llll}
    \toprule
    Model & Top-1 (\%) & Top-3 (\%) & Top-5 (\%) \\
    \midrule
    ResNet-50           & 80.72 ± 0.86 & 92.96 ± 1.17 & 95.64 ± 1.16 \\
    EfficientNet-B3     & 77.08 ± 0.56 & 91.26 ± 1.16 & 94.77 ± 0.96 \\
    MobileNet-V3-Large  & 83.64 ± 1.12 & 95.01 ± 0.68 & 97.05 ± 0.47 \\
    DenseNet-121        & 85.65 ± 0.38 & \textbf{95.82 ± 0.56} & \textbf{97.87 ± 0.23} \\
    \textbf{OliveGemma} & \textbf{92.96 ± 0.91} & 95.73 ± 0.38 & 96.10 ± 0.44 \\
    \bottomrule
  \end{tabular}
  \label{tab:model_performance}
\end{table}

The results in Table~\ref{tab:model_performance} reveal a clear ordering among the convolutional baselines. DenseNet-121 emerges as the strongest CNN at top-1 (85.65±0.38\%), followed by MobileNet-V3-Large (83.64±1.12\%), ResNet-50 (80.72±0.86\%) and EfficientNet-B3 (77.08±0.56\%). This ranking does not track raw model capacity or ImageNet performance, where EfficientNet-B3 and ResNet-50 typically lead, which is informative in itself. Because every backbone is frozen and only the linear head is trained, the comparison reflects the quality of each network's pretrained features for fine-grained food discrimination rather than its capacity to learn new representations. The advantage of DenseNet-121 is therefore consistent with its dense connectivity pattern, which promotes feature reuse across layers and tends to preserve the local texture and colour cues that separate visually adjacent dishes. The comparatively weak showing of EfficientNet-B3, despite its strong supervised ImageNet accuracy, suggests that compound-scaled features optimised for generic object categories transfer less readily to the texture-driven, compositional structure of plated food when the backbone cannot be adapted.

A second observation concerns the gap between top-1 and the deeper ranks. All four CNNs recover sharply at top-3 and top-5, with DenseNet-121 reaching 95.82±0.56\% and 97.87±0.23\% respectively. The size of this jump, more than ten percentage points between top-1 and top-3 for every model, indicates that the correct label is usually present within the network's shortlist even when it is not promoted to first place. In other words, the frozen features carry enough signal to narrow each plate to a small candidate set, but the linear head alone cannot resolve the final fine-grained decision among visually similar classes. This pattern motivates the comparison that follows: where the CNN shortlists are well-formed yet the rank-1 decision is unreliable, the language conditioned predictions of OliveGemma are expected to convert that latent discriminative signal into a correct top-1 answer more often. Notably, OliveGemma's top-1 of 92.96±0.91
92.96±0.91\% exceeds the best CNN by 7.31 points, while the CNN baselines retain a marginal edge at top-5, a trade-off examined in the Discussion.

\subsection{Comparison with Gemini, OpenAI and Anthropic Proprietory Family Models}

OliveGemma was further evaluated against several proprietary frontier vision-language models from the Google (Gemini), OpenAI, and Anthropic model families. All proprietary models were evaluated in the zero-shot setting without any task-specific fine-tuning on the employed food datasets. To ensure a consistent comparison, every model was evaluated using the identical 3 fold cross-validation splits employed for OliveGemma and the CNN baselines. Furthermore, all models operated under the same closed-vocabulary protocol based on the canonical set of 216 food categories. The complete label vocabulary was incorporated into the system prompt and models were instructed to return a single line response whose \texttt{candidates} field is a ranked top-$5$ list in which every entry must be copied \emph{verbatim} from the closed set, no invented, translated, split, or recombined labels, and composed multi-component plates (e.g. \texttt{salmon, leek mashed potatoes}) treated as one indivisible label. Decoding is deterministic (temperature $0$), the response budget is $512$ tokens, images are sent at a maximum side of $1{,}024$\,pixels, and Gemini-3, OpenAI and Claude ``thinking'' is disabled (\texttt{thinking\_budget}=0) so that no reasoning tokens leak into or truncate the structured output. Figure~\ref{fig:gemini_prompt} presents the complete system prompt used throughout the evaluation. Performance was assessed using the same top-1, top-3, and top-5 exact-match metrics adopted for all experiments. Moreover, during inference each model receives this system instruction together with the food image and a recognition question (e.g. \emph{``What is the name of this dish?''}), and the ranked \texttt{candidates} list is scored with the same top-$1$/top-$3$/top-$5$ exact match used throughout. This protocol transformed otherwise open-vocabulary vision-language models into a directly comparable closed-vocabulary classification setting. Table~\ref{tab:gemini} summarizes the average performance across the three cross-validation folds. OliveGemma achieved a mean top-1 accuracy of $92.96\pm0.91\%$, substantially outperforming all evaluated proprietary models. The strongest proprietary baseline, Gemini 3.5 Flash, achieved $74.80\pm0.45\%$ top-1 accuracy, approximately 18 percentage points below OliveGemma. Similar improvements were observed over Gemini 3 Flash, while the performance gap increased considerably for OpenAI ChatGPT 5.4 Mini and Claude Haiku 4.6. 
\begin{table}[htbp]
  \caption{3-fold dish-recognition comparison between zero-shot
  Gemini, OpenAI and Anthropic family models and OliveGemma, on the identical $216$-class closed
  vocabulary and test splits across folds. All models are scored with top-$1$/top-$3$/top-$5$
  exact match.}
  \centering
  \begin{tabular}{lccc}
    \toprule
    Model & Top-1 (\%) & Top-3 (\%) & Top-5 (\%) \\
    \midrule
    Gemini 3 Flash   &  74.17 ± 0.35 &  89.76 ± 0.16 &   92.76 ± 0.34 \\
    Gemini 3.5 Flash &  74.80 ± 0.45 &   90.43 ± 0.02 &   93.25 ± 0.19 \\
    OpenAI ChatGPT 5.4 Mini & 46.35 ± 1.99 & 65.96 ± 2.27 &  72.30 ± 2.02 \\
    Claude Haiku 4.6 &  28.61 ± 1.26 & 42.29 ± 1.42 & 47.59 ± 1.21 \\
    \textbf{OliveGemma} & \textbf{92.96 ± 0.91} & \textbf{95.73 ± 0.38} & \textbf{96.10 ± 0.44} \\
    \bottomrule
  \end{tabular}
  \label{tab:gemini}
\end{table}
The differences become smaller when considering top-3 and top-5 accuracy. Both Gemini models frequently included the correct food category among their highest-ranked predictions but failed to consistently assign it the highest confidence. This behavior is expected in fine-grained food recognition, where visually similar dishes often differ only by subtle ingredients, preparation methods, or presentation. In contrast, OliveGemma was specifically adapted to this domain through LoRA fine-tuning, enabling more reliable discrimination between closely related Mediterranean and European food categories. The comparatively lower performance of ChatGPT 5.4 Mini and Claude Haiku 4.6 further illustrates that general-purpose vision-language capabilities alone are insufficient for this task. Although these models demonstrate strong general visual understanding, they were not optimized for distinguishing highly similar food classes within a specialized closed vocabulary. These findings highlight the importance of domain adaptation and indicate that parameter-efficient fine-tuning can enable relatively small open-weight models to outperform substantially larger proprietary systems on specialized food recognition tasks.

\begin{figure}[hbt]
  \centering
  \begin{minipage}{0.95\linewidth}
  \footnotesize
\begin{verbatim}
[SYSTEM INSTRUCTION -- sent with every dish_name call]

You are a careful food-recognition expert. You will see one image of a
plated dish and one question about it. Answer concisely and exactly in
the requested format.

TASK: identify the dish in the image by choosing from the CLOSED label
set below.

OUTPUT - follow EXACTLY:
- Output ONE single line of JSON and NOTHING else.
- Do NOT think out loud, do NOT explain, do NOT add any text, notes, or
  markdown fences before or after the JSON.
- The JSON must be this exact schema:
  {"candidates": ["<best>", "<2nd>", "<3rd>", "<4th>", "<5th>"]}
- `candidates` is your ranked top-5: exactly 5 labels ordered from MOST
  to LEAST likely (use fewer only if the set has fewer than 5 labels).
- The FIRST element is your single best answer.
- Every element MUST be copied VERBATIM from the closed set below - same
  lowercase, same spelling, same punctuation. Do NOT invent, translate,
  split, or recombine labels. A label that reads as a comma-separated
  plate (e.g. 'salmon, leek mashed potatoes') is ONE indivisible label:
  copy it whole.
- No duplicates in the list.

CLOSED DISH LABEL SET (216 labels, lowercase):
  - almond cream cake
  - almonds
  - anthotyro
  - ...  [ 213 further canonical labels ]  ...
  - zucchini fritters

[USER CONTENT] : <food image>  +  <question, e.g. "What is the name
                                                  of this dish?">
\end{verbatim}
  \end{minipage}
  \caption{System prompt and inputs given to Gemini, OpenAI and Anthropic models. The complete
  $216$-class closed vocabulary is appended to the system instruction while the model must return a ranked top-$5$ of labels copied
  verbatim from this set. The same closed-vocabulary protocol is applied to all
  frontier models for fairness.}
  \label{fig:gemini_prompt}
\end{figure}

\subsection{Likely Ingredients Reasoning}

Beyond closed-vocabulary dish recognition, OliveGemma is trained to enumerate the \emph{likely ingredients} of a plate, an auxiliary attribute that probes whether the model has acquired genuine food understanding rather than a surface image-to-label mapping. Each reference answer, distilled from the teacher model, is a structured list split into a \emph{visible-typical} set (ingredients that should be discernible in the image) and a \emph{commonly-present but not visually confirmable} set (ingredients implied by the dish but typically hidden, e.g. salt, yeast, or egg yolks). Because the answer is an unordered set rather than a single token, exact string match is uninformative. Therefore every prediction and reference is parsed into ingredient sets and grade the prediction against the reference with sample-averaged set-overlap metrics: precision, recall, $F_1$, the Jaccard index, and the stricter exact-set rate (the fraction of images for which the predicted set matches the reference set \emph{exactly}). Table~\ref{tab:likely_ing} reports the results where, OliveGemma recovers the reference ingredient set with an $F_1$ of $95.5\pm0.6\%$ and reproduces the entire set verbatim for $90.8\pm1.3\%$ of images, with precision ($95.7\%$) and recall ($95.3\%$) closely balanced as the model neither systematically over nor under-generates ingredients.

\begin{table}[htbp]
  \caption{Likely-ingredient reasoning of OliveGemma, graded as set overlap
  between the predicted and reference ingredient lists. Values are
  sample-averaged percentages per fold; the last row is the mean $\pm$ standard
  deviation across the three folds. \textit{Exact-set} is the fraction of images
  whose predicted ingredient set matches the reference exactly.}
  \centering
  \begin{tabular}{lcccccc}
    \toprule
    Fold & Precision & Recall & $F_1$ & Jaccard & Exact-set \\
    \midrule
    0 &  96.29 & 95.97 & 96.10 & 95.36 & 91.98 \\
    1 & 95.61 & 95.08 & 95.32 & 94.50 & 90.98 \\
    2 & 95.22 & 94.82 & 94.99 & 94.01 & 89.41 \\
    \midrule
    \textbf{Mean $\pm$ std} & $95.71{\pm}0.54$ & $95.29{\pm}0.60$ & $\mathbf{95.47{\pm}0.57}$ & $94.62{\pm}0.68$ & $90.79{\pm}1.30$ \\
    \bottomrule
  \end{tabular}
  \label{tab:likely_ing}
\end{table}

Table~\ref{tab:likely_ing_examples} illustrates the behaviour qualitatively. For canonical, well-represented dishes the model is frequently verbatim-correct: for \emph{tiramisu} it reproduces both the visible components (cocoa, mascarpone, ladyfingers, coffee-soaked sponge) and the hidden ones (espresso, sugar, egg yolks, Marsala) exactly. The partial cases are revealing as on the bacon-ham-pepper \emph{pizza} the model substitutes \texttt{sliced mushrooms} for one of the peppers which is a visually plausible topping confusion while it recovers the crust, cheese, cured meats, and the hidden leavening agents. Most errors are of this substitution type rather than wholesale hallucination, and the \emph{commonly-present} (hidden) set is recovered almost perfectly even when the visible set drifts, as for the \emph{green apple} where the model misreads the skin colour but still lists the correct fructose/malic-acid/fibre chemistry. The clearest failure mode is downstream of recognition whereas, when the dish itself is misjudged, i.e. \emph{russian salad} read as a leafy green salad, the entire visible ingredient set follows the wrong dish, confirming that ingredient quality is bounded by recognition quality. Overall, these results indicate that the LoRA adaptation instils a structured, decompositional understanding of food rather than a flat label map.

\begin{table}[htb]
  \caption{Representative \texttt{likely\_ingredients} predictions versus the
  distilled reference. ``Visible'' abbreviates \emph{visible-typical ingredients}
  and ``Hidden'' the \emph{commonly-present but not visually confirmable} set.
  $F_1$ is the per-image set-overlap score.}
  \centering
  \footnotesize
  \setlength{\tabcolsep}{4pt}
  \begin{tabular}{p{1.4cm}p{5.3cm}p{5.3cm}c}
    \toprule
    Dish & Reference & OliveGemma prediction & $F_1$ \\
    \midrule
    Tiramisu &
    \textit{Visible:} cocoa powder, mascarpone cream, ladyfinger biscuits, coffee-soaked sponge. \textit{Hidden:} espresso, sugar, egg yolks, Marsala wine. &
    \textit{Visible:} cocoa powder, mascarpone cream, ladyfinger biscuits, coffee-soaked sponge. \textit{Hidden:} espresso, sugar, egg yolks, Marsala wine. &
    1.00 \\
    \addlinespace
    Pizza (bacon, ham, peppers) &
    \textit{Visible:} bacon bits, sliced ham, green bell peppers, mozzarella cheese, tomato sauce, pizza crust. \textit{Hidden:} olive oil, salt, yeast, sugar. &
    \textit{Visible:} bacon bits, sliced ham, bell peppers, \underline{sliced mushrooms}, mozzarella cheese, tomato sauce. \textit{Hidden:} pizza dough, olive oil, yeast, salt. &
    0.70 \\
    \addlinespace
    Green apple &
    \textit{Visible:} green skin, stem, flesh, seeds. \textit{Hidden:} fructose, malic acid, water, fiber. &
    \textit{Visible:} \underline{yellow skin}, green stem, smooth waxy surface, fibrous interior. \textit{Hidden:} fructose, malic acid, water, fiber. &
    0.50 \\
    \addlinespace
    Russian salad &
    \textit{Visible:} diced potatoes, green peas, carrots, boiled eggs, pickles, mayonnaise. \textit{Hidden:} salt, black pepper, mustard, vinegar. &
    \textit{Visible:} \underline{lettuce leaves, cherry tomatoes}, lemon wedges, \underline{
    cucumber slices}, red onion strips, shredded carrots. \textit{Hidden:} olive oil, vinegar, salt, black pepper. &
    0.30 \\
    \bottomrule
  \end{tabular}
  \label{tab:likely_ing_examples}
\end{table}

\section{Discussion}

This study presented OliveGemma, a domain-adapted version of PaliGemma-2-3B for fine-grained food recognition and visual food understanding. The proposed model was trained on a unified dataset comprising the MedGR, ODIN, and VIPPSTAR datasets, resulting in a canonical vocabulary of 216 Mediterranean and European food categories. Beyond food classification, OliveGemma was instruction tuned to perform complementary reasoning tasks, including ingredient identification, visual evidence generation, and discrimination between visually similar dishes. Experimental evaluation demonstrated that the proposed model consistently outperformed all evaluated CNN baselines as well as several contemporary proprietary vision-language models operating under the same closed-vocabulary evaluation protocol.

The performance of OliveGemma should be interpreted in the context of both previous food recognition research and the evaluation setting adopted in this study. Conventional CNN based approaches evaluated on Food-101 \cite{bossard2014food101} typically report top-1 accuracies from 85\%-92\%, while comparable performance has been reported on larger benchmarks such as Food2K \cite{min2023food2k}. However, these benchmarks primarily contain single-label dishes and comparatively homogeneous label spaces. In contrast, the task addressed here involves 216 multilingual food categories originating from three heterogeneous European datasets, introducing substantially greater visual variability, cultural diversity, and class imbalance. Within this considerably more challenging setting, OliveGemma achieved a mean top-1 accuracy of $92.96 \pm 0.91\%$, demonstrating that parameter-efficient adaptation of a general-purpose vision-language model can achieve performance comparable to state-of-the-art CNNs while operating on a substantially richer and more complex label space.

The comparison with the CNN baselines further illustrates the advantages of multimodal instruction tuning. All CNNs were evaluated under an identical evaluation scheme, where only the classification layer was optimized while the pretrained ImageNet feature extractor remained frozen. This design intentionally isolated the representational capabilities of each architecture from differences in optimization strategy. OliveGemma exceeded the strongest CNN baseline, DenseNet-121, by 7.31 percentage points in top-1 accuracy, although DenseNet-121 achieved marginally higher top-3 and top-5 accuracy. This observation suggests that CNN feature extractors often rank the correct class among their highest-confidence predictions but are less effective at selecting the correct class as the most probable output. In contrast, OliveGemma benefits from jointly modelling visual information and language representations, allowing predictions to be conditioned not only on image features but also on semantic relationships encoded during instruction tuning. Similar observations have been reported in recent studies on instruction tuned vision-language models \cite{liu2023llava,steiner2024paligemma2}.

The comparison with frontier vision-language models highlights the importance of domain adaptation. Although Gemini Flash 3, Gemini Flash 3.5, GPT-5.4 Mini, and Claude Haiku 4.6 are substantially larger and more general-purpose systems, they were consistently outperformed by OliveGemma under the closed-vocabulary evaluation protocol. The relatively small differences observed in top-3 and top-5 accuracy indicate that these models frequently identify the correct food among their highest-ranked predictions but struggle to distinguish visually similar food categories when a single prediction is required. These findings suggest that model scale alone does not guarantee superior performance in specialized domains. Instead, domain-specific adaptation through parameter-efficient fine-tuning remains essential for tasks involving subtle visual distinctions, multilingual label spaces, and culturally specific food categories. This observation is consistent with previous work demonstrating the effectiveness of LoRA for adapting large multimodal foundation models to specialized application domains \cite{hu2022lora,li2023llavamed,kuckreja2024geochat}.

Beyond recognition performance, OliveGemma offers several practical advantages. Unlike proprietary frontier models that are accessible only through cloud-based APIs, OliveGemma is fully reproducible, openly available, and can be deployed entirely on local infrastructure. This eliminates recurring inference costs while improving reproducibility and enabling inference without network connectivity. These characteristics are particularly important for clinical dietary assessment, where food images may contain patient-identifiable information and are subject to privacy regulations such as GDPR. The ability to perform inference entirely within institutional infrastructure makes OliveGemma a practical alternative for healthcare environments and contributes to the growing evidence that relatively small, open-weight foundation models can effectively address specialized biomedical tasks when combined with parameter-efficient adaptation.

Aside quantitative improvements, these findings have important implications for automated dietary assessment. The improved performance of OliveGemma suggests that accurate food recognition benefits not only from visual feature extraction but also from the semantic knowledge acquired during multimodal pretraining. Unlike conventional CNN-based classifiers, which learn fixed mappings between images and predefined classes, the vision-language model can exploit relationships between visual appearance and textual concepts, enabling better discrimination of visually similar dishes. Furthermore, the strong performance achieved through parameter-efficient LoRA fine-tuning indicates that adapting a pretrained foundation model to a specialized food domain is more effective than relying solely on increasingly larger general-purpose models. This observation reinforces the idea that domain adaptation, rather than model scale alone, is an important factor for achieving robust performance in complex food recognition tasks. Furthermore, it is important of AI systems to combine high recognition accuracy with explainability, reproducibility, and privacy-preserving deployment for dietary assessment applications. Our findings demonstrate that an open-weight vision-language model can satisfy these requirements simultaneously, outperforming both conventional CNN architectures and several proprietary frontier VLMs while remaining suitable for local deployment. Consequently, OliveGemma provides a practical foundation for future clinical and nutritional applications, where reliable food recognition can improve downstream tasks such as nutritional analysis, dietary monitoring, and decision-support systems without requiring dependence on proprietary cloud-based services.

The present study nevertheless has several limitations. The unified dataset covers 216 Mediterranean and European food categories, yet many regional cuisines remain unrepresented. Expanding the training corpus with additional datasets, such as the Central Asian Food Dataset \cite{centralasian}, would increase both the diversity and complexity of the label space, particularly for visually similar carbohydrate-based dishes including rice and noodle varieties. Future work will also investigate extending OliveGemma beyond closed-vocabulary recognition toward open-vocabulary food understanding, nutritional estimation, and integration with clinical dietary assessment systems.

Overall, the findings demonstrate that parameter-efficient adaptation of an open-weight vision-language model provides an effective solution for fine-grained food recognition. By combining multimodal instruction tuning with LoRA, OliveGemma achieves competitive recognition performance while providing explainable food understanding, local deployment, and reproducible experimentation.

\section{Disclosure on AI Use}
In the current study we have employed Anthropic Claude Opus for english language refinements and OpenAI ChatGPT for Figure 1 stylistic modifications while this figure derived from originally designed graphics by the authors. All the english language and figures modifications were reviewed by the authors of this study to avoid hallucinated artifacts.

\section{Acknowledgments}
This work is supported by the Vippstar project, funded by the European Union’s Horizon 2020 research and innovation program under grant agreement No. 101156763. It reflects only the author’s view. The Commission is not responsible for any use that may be made of the information it contains.

\bibliographystyle{unsrt}  
\bibliography{references}  

@article{estruch2018predimed,
  title   = {Primary Prevention of Cardiovascular Disease with a {Mediterranean}
             Diet Supplemented with Extra-Virgin Olive Oil or Nuts},
  author  = {Estruch, Ram{\'o}n and Ros, Emilio and Salas-Salvad{\'o}, Jordi and
             Covas, Maria-Isabel and Corella, Dolores and Ar{\'o}s, Fernando and
             G{\'o}mez-Gracia, Enrique and Ruiz-Guti{\'e}rrez, Valentina and
             Fiol, Miquel and Lapetra, Jos{\'e} and Lamuela-Raventos, Rosa Maria
             and Serra-Majem, Llu{\'i}s and Pint{\'o}, Xavier and Basora, Josep
             and Mu{\~n}oz, Miguel Angel and Sorl{\'i}, Jos{\'e} V. and
             Mart{\'i}nez, Jos{\'e} Alfredo and Mart{\'i}nez-Gonz{\'a}lez,
             Miguel Angel},
  journal = {New England Journal of Medicine},
  volume  = {378},
  number  = {25},
  pages   = {e34},
  year    = {2018},
  doi     = {10.1056/NEJMoa1800389}
}

@article{wang2022dietary,
  title   = {A review on vision-based analysis for automatic dietary assessment},
  author  = {Wang, Wei and Min, Weiqing and Li, Tianhao and Dong, Xiaoxiao and
             Li, Haisheng and Jiang, Shuqiang},
  journal = {Trends in Food Science \& Technology},
  volume  = {122},
  pages   = {223--237},
  year    = {2022},
  doi     = {10.1016/j.tifs.2022.02.017}
}

@article{min2019survey,
  title   = {A Survey on Food Computing},
  author  = {Min, Weiqing and Jiang, Shuqiang and Liu, Linhu and Rui, Yong and
             Jain, Ramesh},
  journal = {ACM Computing Surveys},
  volume  = {52},
  number  = {5},
  pages   = {1--36},
  articleno = {92},
  year    = {2019},
  doi     = {10.1145/3329168}
}

@inproceedings{bossard2014food101,
  title     = {Food-101 -- Mining Discriminative Components with Random Forests},
  author    = {Bossard, Lukas and Guillaumin, Matthieu and Van Gool, Luc},
  booktitle = {European Conference on Computer Vision (ECCV)},
  series    = {Lecture Notes in Computer Science},
  volume    = {8694},
  pages     = {446--461},
  year      = {2014},
  publisher = {Springer},
  doi       = {10.1007/978-3-319-10599-4_29}
}

@article{min2023food2k,
  title   = {Large Scale Visual Food Recognition},
  author  = {Min, Weiqing and Wang, Zhiling and Liu, Yuxin and Luo, Mengjiang and
             Kang, Liping and Wei, Xiaoming and Wei, Xiaolin and Jiang, Shuqiang},
  journal = {IEEE Transactions on Pattern Analysis and Machine Intelligence},
  volume  = {45},
  number  = {8},
  pages   = {9932--9949},
  year    = {2023},
  doi     = {10.1109/TPAMI.2023.3237871}
}

@inproceedings{radford2021clip,
  title     = {Learning Transferable Visual Models From Natural Language
               Supervision},
  author    = {Radford, Alec and Kim, Jong Wook and Hallacy, Chris and Ramesh,
               Aditya and Goh, Gabriel and Agarwal, Sandhini and Sastry, Girish
               and Askell, Amanda and Mishkin, Pamela and Clark, Jack and
               Krueger, Gretchen and Sutskever, Ilya},
  booktitle = {Proceedings of the 38th International Conference on Machine
               Learning (ICML)},
  series    = {Proceedings of Machine Learning Research},
  volume    = {139},
  pages     = {8748--8763},
  year      = {2021}
}

@inproceedings{zhai2023siglip,
  title     = {Sigmoid Loss for Language Image Pre-Training},
  author    = {Zhai, Xiaohua and Mustafa, Basil and Kolesnikov, Alexander and
               Beyer, Lucas},
  booktitle = {Proceedings of the IEEE/CVF International Conference on Computer
               Vision (ICCV)},
  pages     = {11975--11986},
  year      = {2023},
  doi       = {10.1109/ICCV51070.2023.01100}
}

@inproceedings{li2023blip2,
  title     = {{BLIP-2}: Bootstrapping Language-Image Pre-training with Frozen
               Image Encoders and Large Language Models},
  author    = {Li, Junnan and Li, Dongxu and Savarese, Silvio and Hoi, Steven},
  booktitle = {Proceedings of the 40th International Conference on Machine
               Learning (ICML)},
  year      = {2023},
  note      = {arXiv:2301.12597}
}

@inproceedings{liu2023llava,
  title     = {Visual Instruction Tuning},
  author    = {Liu, Haotian and Li, Chunyuan and Wu, Qingyang and Lee, Yong Jae},
  booktitle = {Advances in Neural Information Processing Systems (NeurIPS)},
  volume    = {36},
  year      = {2023},
  note      = {arXiv:2304.08485}
}

@article{bai2025qwenvl,
  title   = {{Qwen2.5-VL} Technical Report},
  author  = {Bai, Shuai and Chen, Keqin and Liu, Xuejing and Wang, Jialin and
             Ge, Wenbin and Song, Sibo and Dang, Kai and Wang, Peng and Wang,
             Shijie and Tang, Jun and Zhong, Humen and Zhu, Yuanzhi and Yang,
             Mingkun and Li, Zhaohai and Wan, Jianqiang and Wang, Pengfei and
             Ding, Wei and Fu, Zheren and Xu, Yiheng and Ye, Jiabo and Zhang,
             Xi and Xie, Tianbao and Cheng, Zesen and Zhang, Hang and Yang,
             Zhibo and Xu, Haiyang and Lin, Junyang},
  journal = {arXiv preprint arXiv:2502.13923},
  year    = {2025}
}

@article{steiner2024paligemma2,
  title   = {{PaliGemma 2}: A Family of Versatile {VLMs} for Transfer},
  author  = {Steiner, Andreas and Pinto, Andr{\'e} Susano and Tschannen, Michael
             and Keysers, Daniel and Wang, Xiao and Bitton, Yonatan and
             Gritsenko, Alexey and Minderer, Matthias and Sherbondy, Anthony and
             Long, Shangbang and Qin, Siyang and Ingle, Reeve and Bugliarello,
             Emanuele and Kazemzadeh, Sahar and Mesnard, Thomas and
             Alabdulmohsin, Ibrahim and Beyer, Lucas and Zhai, Xiaohua},
  journal = {arXiv preprint arXiv:2412.03555},
  year    = {2024}
}

@inproceedings{hu2022lora,
  title     = {{LoRA}: Low-Rank Adaptation of Large Language Models},
  author    = {Hu, Edward J. and Shen, Yelong and Wallis, Phillip and
               Allen-Zhu, Zeyuan and Li, Yuanzhi and Wang, Shean and Wang, Lu and
               Chen, Weizhu},
  booktitle = {International Conference on Learning Representations (ICLR)},
  year      = {2022},
  note      = {arXiv:2106.09685}
}

@inproceedings{dettmers2023qlora,
  title     = {{QLoRA}: Efficient Finetuning of Quantized {LLMs}},
  author    = {Dettmers, Tim and Pagnoni, Artidoro and Holtzman, Ari and
               Zettlemoyer, Luke},
  booktitle = {Advances in Neural Information Processing Systems (NeurIPS)},
  volume    = {36},
  year      = {2023},
  note      = {arXiv:2305.14314}
}

@article{kirkpatrick2017overcoming,
  title   = {Overcoming catastrophic forgetting in neural networks},
  author  = {Kirkpatrick, James and Pascanu, Razvan and Rabinowitz, Neil and
             Veness, Joel and Desjardins, Guillaume and Rusu, Andrei A. and
             Milan, Kieran and Quan, John and Ramalho, Tiago and
             Grabska-Barwinska, Agnieszka and Hassabis, Demis and Clopath,
             Claudia and Kumaran, Dharshan and Hadsell, Raia},
  journal = {Proceedings of the National Academy of Sciences},
  volume  = {114},
  number  = {13},
  pages   = {3521--3526},
  year    = {2017},
  doi     = {10.1073/pnas.1611835114}
}

@inproceedings{kawano2014uecfood256,
  title     = {Automatic Expansion of a Food Image Dataset Leveraging Existing
               Categories with Domain Adaptation},
  author    = {Kawano, Yoshiyuki and Yanai, Keiji},
  booktitle = {European Conference on Computer Vision (ECCV) Workshops --
               Transferring and Adapting Source Knowledge in Computer Vision
               (TASK-CV)},
  pages     = {3--17},
  year      = {2014},
  publisher = {Springer}
}

@inproceedings{min2020isia,
  title     = {{ISIA Food-500}: A Dataset for Large-Scale Food Recognition via
               Stacked Global-Local Attention Network},
  author    = {Min, Weiqing and Liu, Linhu and Wang, Zhiling and Luo, Zhengdong
               and Wei, Xiaoming and Wei, Xiaolin and Jiang, Shuqiang},
  booktitle = {Proceedings of the 28th ACM International Conference on
               Multimedia (MM)},
  pages     = {393--401},
  year      = {2020},
  doi       = {10.1145/3394171.3414031}
}

@inproceedings{salvador2017recipe1m,
  title     = {Learning Cross-Modal Embeddings for Cooking Recipes and Food
               Images},
  author    = {Salvador, Amaia and Hynes, Nicholas and Aytar, Yusuf and Marin,
               Javier and Ofli, Ferda and Weber, Ingmar and Torralba, Antonio},
  booktitle = {Proceedings of the IEEE Conference on Computer Vision and Pattern
               Recognition (CVPR)},
  pages     = {3020--3028},
  year      = {2017},
  doi       = {10.1109/CVPR.2017.327}
}

@inproceedings{he2016resnet,
  title     = {Deep Residual Learning for Image Recognition},
  author    = {He, Kaiming and Zhang, Xiangyu and Ren, Shaoqing and Sun, Jian},
  booktitle = {Proceedings of the IEEE Conference on Computer Vision and Pattern
               Recognition (CVPR)},
  pages     = {770--778},
  year      = {2016},
  doi       = {10.1109/CVPR.2016.90}
}

@inproceedings{szegedy2015googlenet,
  title     = {Going Deeper with Convolutions},
  author    = {Szegedy, Christian and Liu, Wei and Jia, Yangqing and Sermanet,
               Pierre and Reed, Scott and Anguelov, Dragomir and Erhan, Dumitru
               and Vanhoucke, Vincent and Rabinovich, Andrew},
  booktitle = {Proceedings of the IEEE Conference on Computer Vision and Pattern
               Recognition (CVPR)},
  pages     = {1--9},
  year      = {2015},
  doi       = {10.1109/CVPR.2015.7298594}
}

@inproceedings{huang2017densenet,
  title     = {Densely Connected Convolutional Networks},
  author    = {Huang, Gao and Liu, Zhuang and van der Maaten, Laurens and
               Weinberger, Kilian Q.},
  booktitle = {Proceedings of the IEEE Conference on Computer Vision and Pattern
               Recognition (CVPR)},
  pages     = {4700--4708},
  year      = {2017},
  doi       = {10.1109/CVPR.2017.243}
}

@inproceedings{tan2019efficientnet,
  title     = {{EfficientNet}: Rethinking Model Scaling for Convolutional Neural
               Networks},
  author    = {Tan, Mingxing and Le, Quoc V.},
  booktitle = {Proceedings of the 36th International Conference on Machine
               Learning (ICML)},
  series    = {Proceedings of Machine Learning Research},
  volume    = {97},
  pages     = {6105--6114},
  year      = {2019}
}

@inproceedings{dosovitskiy2021vit,
  title     = {An Image is Worth 16x16 Words: Transformers for Image Recognition
               at Scale},
  author    = {Dosovitskiy, Alexey and Beyer, Lucas and Kolesnikov, Alexander and
               Weissenborn, Dirk and Zhai, Xiaohua and Unterthiner, Thomas and
               Dehghani, Mostafa and Minderer, Matthias and Heigold, Georg and
               Gelly, Sylvain and Uszkoreit, Jakob and Houlsby, Neil},
  booktitle = {International Conference on Learning Representations (ICLR)},
  year      = {2021},
  note      = {arXiv:2010.11929}
}

@inproceedings{lin2015bilinear,
  title     = {Bilinear {CNN} Models for Fine-Grained Visual Recognition},
  author    = {Lin, Tsung-Yu and RoyChowdhury, Aruni and Maji, Subhransu},
  booktitle = {Proceedings of the IEEE International Conference on Computer
               Vision (ICCV)},
  pages     = {1449--1457},
  year      = {2015},
  doi       = {10.1109/ICCV.2015.170}
}

@article{bai2023qwenvl,
  title   = {{Qwen-VL}: A Versatile Vision-Language Model for Understanding,
             Localization, Text Reading, and Beyond},
  author  = {Bai, Jinze and Bai, Shuai and Yang, Shusheng and Wang, Shijie and
             Tan, Sinan and Wang, Peng and Lin, Junyang and Zhou, Chang and
             Zhou, Jingren},
  journal = {arXiv preprint arXiv:2308.12966},
  year    = {2023}
}

@article{beyer2024paligemma,
  title   = {{PaliGemma}: A Versatile 3B {VLM} for Transfer},
  author  = {Beyer, Lucas and Steiner, Andreas and Pinto, Andr{\'e} Susano and
             Kolesnikov, Alexander and Wang, Xiao and Salz, Daniel and Neumann,
             Maxim and Alabdulmohsin, Ibrahim and Tschannen, Michael and
             Bugliarello, Emanuele and Unterthiner, Thomas and Keysers, Daniel
             and Koppula, Skanda and Liu, Fangyu and Grycner, Adam and
             Gritsenko, Alexey and Houlsby, Neil and Kumar, Manoj and Rong,
             Keran and Eisenschlos, Julian and Kabra, Rishabh and Bauer,
             Matthias and Bo{\v{s}}njak, Matko and Chen, Xi and Minderer,
             Matthias and Voigtlaender, Paul and Bica, Ioana and Balazevic,
             Ivana and Puigcerver, Joan and Papalampidi, Pinelopi and Henaff,
             Olivier and Xiong, Xi and Soricut, Radu and Harmsen, Jeremiah and
             Zhai, Xiaohua},
  journal = {arXiv preprint arXiv:2407.07726},
  year    = {2024}
}

@inproceedings{chen2024internvl,
  title     = {{InternVL}: Scaling up Vision Foundation Models and Aligning for
               Generic Visual-Linguistic Tasks},
  author    = {Chen, Zhe and Wu, Jiannan and Wang, Wenhai and Su, Weijie and
               Chen, Guo and Xing, Sen and Zhong, Muyan and Zhang, Qinglong and
               Zhu, Xizhou and Lu, Lewei and Li, Bin and Luo, Ping and Lu,
               Tong and Qiao, Yu and Dai, Jifeng},
  booktitle = {Proceedings of the IEEE/CVF Conference on Computer Vision and
               Pattern Recognition (CVPR)},
  pages     = {24185--24198},
  year      = {2024}
}

@inproceedings{houlsby2019adapter,
  title     = {Parameter-Efficient Transfer Learning for {NLP}},
  author    = {Houlsby, Neil and Giurgiu, Andrei and Jastrzebski, Stanislaw and
               Morrone, Bruna and de Laroussilhe, Quentin and Gesmundo, Andrea
               and Attariyan, Mona and Gelly, Sylvain},
  booktitle = {Proceedings of the 36th International Conference on Machine
               Learning (ICML)},
  series    = {Proceedings of Machine Learning Research},
  volume    = {97},
  pages     = {2790--2799},
  year      = {2019}
}

@inproceedings{li2021prefix,
  title     = {Prefix-Tuning: Optimizing Continuous Prompts for Generation},
  author    = {Li, Xiang Lisa and Liang, Percy},
  booktitle = {Proceedings of the 59th Annual Meeting of the Association for
               Computational Linguistics and the 11th International Joint
               Conference on Natural Language Processing (ACL-IJCNLP)},
  pages     = {4582--4597},
  year      = {2021},
  doi       = {10.18653/v1/2021.acl-long.353}
}

@inproceedings{lester2021prompt,
  title     = {The Power of Scale for Parameter-Efficient Prompt Tuning},
  author    = {Lester, Brian and Al-Rfou, Rami and Constant, Noah},
  booktitle = {Proceedings of the 2021 Conference on Empirical Methods in
               Natural Language Processing (EMNLP)},
  pages     = {3045--3059},
  year      = {2021},
  doi       = {10.18653/v1/2021.emnlp-main.243}
}

@inproceedings{li2023llavamed,
  title     = {{LLaVA-Med}: Training a Large Language-and-Vision Assistant for
               Biomedicine in One Day},
  author    = {Li, Chunyuan and Wong, Cliff and Zhang, Sheng and Usuyama, Naoto
               and Liu, Haotian and Yang, Jianwei and Naumann, Tristan and Poon,
               Hoifung and Gao, Jianfeng},
  booktitle = {Advances in Neural Information Processing Systems (NeurIPS),
               Datasets and Benchmarks Track},
  volume    = {36},
  year      = {2023},
  note      = {arXiv:2306.00890}
}

@inproceedings{kuckreja2024geochat,
  title     = {{GeoChat}: Grounded Large Vision-Language Model for Remote
               Sensing},
  author    = {Kuckreja, Kartik and Danish, Muhammad Sohail and Naseer, Muzammal
               and Das, Abhijit and Khan, Salman and Khan, Fahad Shahbaz},
  booktitle = {Proceedings of the IEEE/CVF Conference on Computer Vision and
               Pattern Recognition (CVPR)},
  pages     = {27831--27840},
  year      = {2024}
}

@inproceedings{howard2019mobilenetv3,
  title     = {Searching for {MobileNetV3}},
  author    = {Howard, Andrew and Sandler, Mark and Chu, Grace and
               Chen, Liang-Chieh and Chen, Bo and Tan, Mingxing and
               Wang, Weijun and Zhu, Yukun and Pang, Ruoming and
               Vasudevan, Vijay and Le, Quoc V. and Adam, Hartwig},
  booktitle = {Proceedings of the IEEE/CVF International Conference on Computer Vision (ICCV)},
  pages     = {1314--1324},
  year      = {2019}
}

@Article{centralasian,
AUTHOR = {Karabay, Aknur and Bolatov, Arman and Varol, Huseyin Atakan and Chan, Mei-Yen},
TITLE = {A Central Asian Food Dataset for Personalized Dietary Interventions},
JOURNAL = {Nutrients},
VOLUME = {15},
YEAR = {2023},
NUMBER = {7},
ARTICLE-NUMBER = {1728},
URL = {https://www.mdpi.com/2072-6643/15/7/1728},
ISSN = {2072-6643},
}

@ARTICLE{medgr,
  title    = "An Automated {Image-Based} Dietary Assessment System for
              Mediterranean Foods",
  author   = "Konstantakopoulos, Fotios S and Georga, Eleni I and Fotiadis,
              Dimitrios I",
  journal  = "IEEE Open J Eng Med Biol",
  volume   =  4,
  pages    = "45--54",
  month    =  apr,
  year     =  2023,
  address  = "United States",
  language = "en"
}

\newpage

\appendix
\section{Appendix I - Food Classes}
Here are presented the exact 216 food classes, where the models have been validated on.

\begin{table}[htbp]
    \centering
    \caption{List of 216 Classes}
    \begin{tabular}{|p{\textwidth}|}
        \hline
        almond cream cake, almonds, anthotyro, apple compote, apricot, asparagus, avocado, baked anchovies, baklava, banana, beef burger, beef in tomato sauce, beef stew, beef stew with celery, beef stew with onions, beef stew french fries, beef with artichokes in egg lemon sauce, beet salad, berries, biscotti, boiled antidia, boiled beef with vegetables, boiled beetroot, boiled cauliflower, boiled chicken, boiled greens, boiled octopus, boiled pork, boiled radikia, boiled stamnagathi, boiled vegetables, boiled vlita, breaded chicken, broccoli, bulbs, cabbage rolls, cabbage salad, calamari with spinach, canned peaches, carbonara, carrot, celery, cheese pie, cherries, chicken in red sauce, chicken souvlaki, chicken croquette applesauce, chicken rice currysauce, chickpea soup, chocolate bundt cake, chocolate cake, codfish plaki, cookies, corn, croissant with cheese, cucumber, cuttlefish with spinach, dolmades, dry fig, egg flan, eggplant salad, feta, fettine di manzo, firikia apples, french croissant, fresh garlic, fresh onion, fried anchovies, fried atherina, fried calamari, fried codfish, fried eggs, fried gopa, fried koutsomoura, fried marida, fried mushrooms, fried mussels, fried peppers, fried potatoes, fried rice, fried safridi, fried zucchini, frittata, galatopita, gelato, giant beans, gnocchi, grapefruit, graviera, greek artichoke stew, green apple, green bean and olive oil stew, green beans, green salad, grilled mushrooms, grilled quince, grilled sardines, hazelnut, honey rolls, imam bayildi, imam bayildi with mince, kalathaki lemnos, kefalograviera, kefalotyri, kiwi, ladotyri, lamb fricassee, lasagna, leek fritters, lemon, lentil soup, lettuce, lettuce salad, loquat, macaroni ham and cheese, mango, manouri, meat pie, meatball tomato sauce mashed potatoes, meatballs, mela, melon, metsovone, mezzo uovo, moustokouloura, nectarines, octopus in vinegar sauce, octopus with macaroni, okra stew, omelette, onion, orange, parsley, pasta bolognese, pasta souffle with cheese, pasta with cheese, pasta with eggplant sauce, pastitsio, peaches, peanuts arapiko, pear, peinirli, penne al ragu, penne al sugo, penne in bianco, pineapple, pineapple compote, pistachio, pizza, pizza margherita, pizza with bacon ham and peppers, plums, pomegranate, pork chop, pork chop fried potatoes chicory, pork in red sauce, potato fritters, potato salad, potatoes in red sauce, prosciutto cotto, prosciutto crudo, pumpkin, pumpkin seeds, purple cabbage, radish, raisins, ravioli, raw quince, raw turnip, red apple, red grapes, rice with pumpkin, risotto, roasted chicken, roasted vegetables, rocket salad, russian salad, salame, salmon leek mashed potatoes, salsiccia, sausage and peppers, sausage with leek, sausage carrot potato, semolina halvah, shrimp saganaki, shrimp salad, sole fillet, spaghetti al ragu, spaghetti al sugo, spaghetti in bianco, spinach, spinach and rice, spinach pie, steak mixed salad, strangolapretti, strawberries, sunflower seeds, sweet semolina cake, tangerine, tas kebab, telemes, tiramisu, tomato, tomato pasta, torta fetta, tuna salad, tzanera, tzatziki, uova strapazzate, walnuts, watermelon, white grapes, yellow pepper, zucchini, zucchini fritters, zucchini pasta. \\ \hline
    \end{tabular}
    \label{tab:food_paragraph}
\end{table}

\section{Appendix II - Accessing \& Inferring the Model}
OliveGemma could be accessed through HuggingFace repository under the following url \url{https://huggingface.co/JamesZar/OliveGemma-3B}. The following snippet presents the commands to download it and enable it locally.

\begin{table}[t]
\centering
\caption{OliveGemma Inference Code}
\label{tab:code_table}
\begin{tabular}{|l|}
\hline
\begin{lstlisting}
import torch
from PIL import Image
from transformers import AutoProcessor, PaliGemmaForConditionalGeneration

REPO = "JamesZar/OliveGemma-3B"

processor = AutoProcessor.from_pretrained(REPO)
model = PaliGemmaForConditionalGeneration.from_pretrained(
    REPO, torch_dtype=torch.bfloat16, device_map="auto"
).eval()

image = Image.open("dish.jpg").convert("RGB")
question = "What is the name of this dish?"

# IMPORTANT: PaliGemma prompt format used in training:
prompt = f"<image>answer en {question}\n"

inputs = processor(text=prompt, images=image, return_tensors="pt").to(model.device)
in_len = inputs["input_ids"].shape[-1]
with torch.no_grad():
    out = model.generate(**inputs, max_new_tokens=64, do_sample=False)
print(processor.decode(out[0][in_len:], skip_special_tokens=True).strip())
\end{lstlisting} \\ \hline
\end{tabular}
\end{table}

\end{document}